# A UNIVERSAL UPDATE-PACING FRAMEWORK FOR VISUAL TRACKING

*Zexi Hu, Yuefang Gao, Dong Wang, Xuhong Tian\**

South China Agricultural University

**ABSTRACT**

This paper proposes a novel framework to alleviate the model drift problem in visual tracking, which is based on paced updates and trajectory selection. Given a base tracker, an ensemble of trackers is generated, in which each tracker's update behavior will be paced and then traces the target object forward and backward to generate a pair of trajectories in an interval. Then, we implicitly perform self-examination based on trajectory pair of each tracker and select the most robust tracker. The proposed framework can effectively leverage temporal context of sequential frames and avoid to learn corrupted information. Extensive experiments on the standard benchmark suggest that the proposed framework achieves superior performance against state-of-the-art trackers.

## 1. INTRODUCTION

Visual tracking is a fundamental and widespread problem in computer vision with numerous applications of interest. The task is to learn an arbitrary target, which is usually an unknown object and a rectangle area of an image, in the first frame and detect its location in the following sequential frames, being called tracking-by-detection. Efforts have been made in past decades and yield significant progress on results' accuracy and robustness [1, 2]. Due to nature of the task, a tracker is required to be generic and used for any kinds of object classes, which is not allowed to apply prior knowledge of any specific classes.

In recent years, several excellent tracking approaches were developed and rely on either discriminative or generative representations. Discriminative approaches consist of training a classifier and predicting image patches to be the target or not, separating them from the background, while proposal image patches come from different parts of a frame, usually surrounding the location of target in the previous frame. In [3], Hare et al. propose a Structured SVM classifier and avoid to rely on a heuristic intermediate step for producing labeled binary samples, which is often a source of error during tracking. In [4], multiple instance learning is used to avoid the error-prone, hard-labeling process. Another discriminative method by Kalal et al. [5] employs a set of structural constraints to guide the sampling process of a boosting classifier. Generative approaches consider the object appearances and search most similar candidates in the current frame. Developed generative methods include holistic templates [6], subspace representations [7].

However, recent trackers have encountered a bottleneck of handling the problem of model drift. Most of the problems are caused by the occurrence of occlusion, fast motion, background clutters and so on. To solve this problem, several methods are developed. A popular way is re-detection. This type of methods is to maintain an additional detector and correct tracker's prediction when error occurs, which can be seen in many trackers, e.g. [5, 8], but the main drawback is the demand of designing a different algorithm to assist the unaware tracker and the increase of computation to maintain such a detector. Another way is to cast out undesired information in update image. Kim et al. [9] propose a feature to decompose current bounding box into ordered patches and assign each one with weighted information, enhancing the importance of object part and alleviating the impact of noise in the bounding box. Possegger et al. [10] propose an efficient discriminative color model to differentiate the target from the background clutters. Some studies employ two or more components to handle information at different occasion. [8, 11–13] are based on this idea, using more than one correlation filters, convolutional neural networks, feature stores or tracker snapshots.

In this paper, we develop a framework which cooperates with existing trackers, to guide them update in proper occasions, utilizing temporal context and alleviating model corruption brought by false updates. We employ an ensemble of trackers, which are initialized from a base tracker, pacing their update behaviors and select the best situation by robustness scores based on trajectories they generated, which is universal to judge trackers' performances. Our framework can be applied universally with most trackers, simply decomposing their processes into tracking and updating. Extensive experiments on a standard visual tracking benchmark [14] with 51 video sequences demonstrate that the proposed framework makes a remarkable improvement on performances of the existing trackers.

\*Corresponding author: tianxuhong@scau.edu.cn (Xuhong Tian). This research is supported by National Natural Science Foundation of China (no. 61202294).

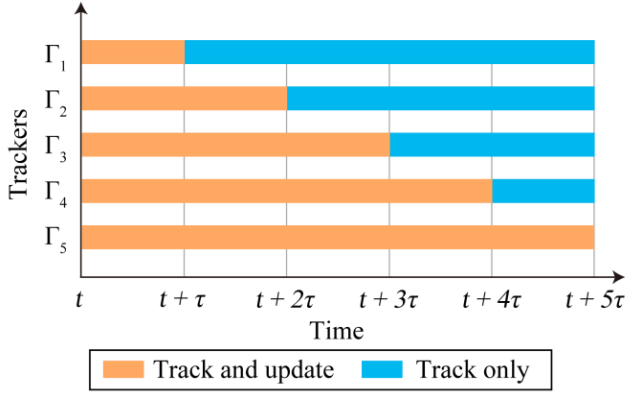

**Fig. 1**. Paced updates: *n* is set as 5 in this diagram, therefore there are 5 trackers and the length of the current process is 5τ.

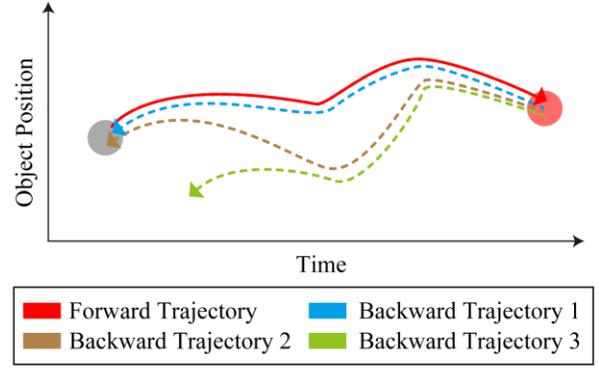

**Fig. 2**. Trajectory analysis: Pairing with forward trajectory, backward trajectory 1 and 2 is cyclic, but 3 is non-cyclic.

## 2. PROPOSED METHOD

A phenomenon has been observed that in a sequence, tracker's model corruption occurs randomly due to change of appearance of the target, which will either recover afterward or remain to form a new appearance. In the first situation, if model keeps updating along the sequence, it may corrupt due to being too adaptive for the temporary change. In the second situation, if model updates conservatively, it may also corrupt for being too less sensitive. Therefore, it's difficult for a tracker to detect its situation. An intuitive solution is to guide tracker to avoid updating in the undesired period with negative information, meanwhile, update in the appropriate occasion. We decompose our solution into two parts, paced updates and trajectory selection.

**Paced Updates:** First of all, we initialize an ensemble **E** of *n* trackers

$$\mathbf{E} = \{\Gamma_1, \Gamma_2, \Gamma_3, \cdots, \Gamma_n\} \quad (1)$$

where $\Gamma_i$ denotes i-th tracker and all of them come from the base tracker $\Gamma_{base}$. Note that even though these *n* trackers are instances from the same tracker, each one will run and behave independently in the following process.

Then, all the trackers begin to track forward from the first frame denoted frame *t* and update their models meanwhile. Until frame *t* + τ, the ensemble completes the first interval, denoted [*t*, *t* + τ]. In the second interval, all the trackers still continue to track forward from frame *t* + τ to *t* + 2τ, but the difference is all the trackers will update their model except $\Gamma_1$, which stops updating from frame *t* + τ and maintains this situation in the following intervals, i.e. $\Gamma_1$ only learns in the interval [*t*, *t* + τ] and discards the knowledge from frame *t* + τ. In a similar manner, $\Gamma_2, \Gamma_3, \ldots, \Gamma_{n-2}, \Gamma_{n-1}$ will stop updating from frame *t* + 2τ, *t* + 3τ, …, *t* + (n-2)τ, *t* + (n-1)τ correspondingly. The last tracker $\Gamma_n$ will have no chance to stop updating since its update process is made to cover all the intervals, which is [*t*, *t* + nτ], i.e. fully updated. Fig. 1 gives an institutive illustration of paced updates. After completing such a process, ensemble **E** can be considered to cover all the possibilities of updating during these intervals.

The trajectory yielded by $\Gamma_i$ from frame $t_1$ to $t_2$, being frame *t* to *t* + *n*τ in the current interval, is denoted by

$$\overrightarrow{\mathbf{X}^i_{t_1:t_2}} = \{\overrightarrow{x^i_{t_1}}, \overrightarrow{x^i_{t_1+1}}, \overrightarrow{x^i_{t_1+2}}, \cdots, \overrightarrow{x^i_{t_2}}\} \quad (2)$$

where $\overrightarrow{x^i_t}$ is the bounding box predicted forward by $\Gamma_i$ at frame *t*.

Then in [$t_1$, $t_2$], trackers in **E** will track backward with an initial bounding box given from the last one in corresponding forward trajectory, and calculate a backward trajectory $\overleftarrow{\mathbf{X}^i_{t_2:t_1}}$, denoted by

$$\overleftarrow{\mathbf{X}^i_{t_2:t_1}} = \{\overleftarrow{x^i_{t_2}}, \overleftarrow{x^i_{t_2-1}}, \overleftarrow{x^i_{t_2-2}}, \cdots, \overleftarrow{x^i_{t_1}}\} \quad (3)$$

where $\overleftarrow{x^i_t}$ is the bounding box predicted backward by $\Gamma_i$ at frame *t*. Hence it's $\overrightarrow{x^i_{t_2}} = \overleftarrow{x^i_{t_2}}$. In backward tracking, trackers are allowed to update in the whole process.

**Trajectory Selection:** Now we obtain *n* pairs of forward and backward trajectories yielded by the tracker ensemble **E**. For every tracker $\Gamma_i$, trajectory analysis between $\overrightarrow{\mathbf{X}^i_{t_1:t_2}}$ and $\overleftarrow{\mathbf{X}^i_{t_2:t_1}}$ is the key to judging whether the tracker succeeds in [$t_1$, $t_2$]. We employ the criterion in [15] to measure the robustness of each tracker in these periods.

The first step is checking of cyclicity. As shown in Fig. 2, pairing with Forward Trajectory, Backward Trajectory 1 and 2 can form a cycle, signifying a high likelihood of successful tracking. However, Backward Trajectory 3 is non-cyclic with Forward Trajectory, implying a large chance of unreliable tracking. Additionally, Backward Trajectory 1 matches Forward Trajectory more accurately owing to the smaller distance between $\overrightarrow{x_t}$ and $\overleftarrow{x_t}$ comparing with Backward Trajectory 2, suggesting to be most reliable among all of these backward trajectories.

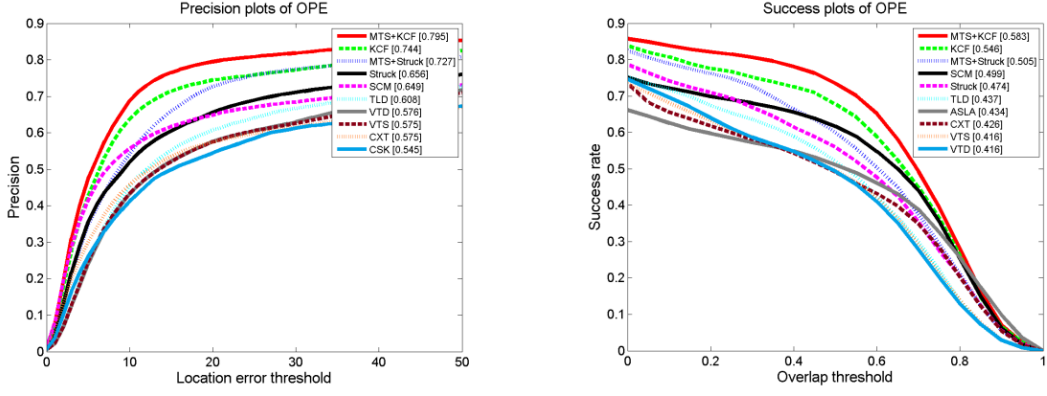

**Fig. 3**. Average precision rate plots (PR, left) and success rate plot (SR, right) of OPE. Style and color of the lines are determined by the rankings, instead of trackers' name.

After checking of cyclicity, geometric similarity and appearance similarity are taken into account. At frame $t$, the geometric similarity is defined as

$$\varsigma_t = \exp\left(-\frac{\|\vec{x_t} - \vec{\hat{x}_t}\|^2}{\sigma_1^2}\right) \quad (4)$$

and the appearance similarity is defined as

$$\phi_t = \exp\left(-\frac{\sum_{Q\in S}\| K \bullet (P(\vec{x_t}) - Q)\|^2}{4wh\sigma_2^2}\right) \quad (5)$$

where $K$ is a Gaussian weight mask, "•" is the pixel-wise weight multiplication, $P(\mathbf{x})$ is image patch of the bounding box $\mathbf{x}$, $S$ is a set of image patches selected as target appearance, usually including the object image in the first frame, $w$ and $h$ are the width and height of a bounding box, respectively.

Finally, combining (4) and (5), the robustness score $\psi$ of a tracker during $[t_1, t_2]$ is obtained, defined as

$$\psi_{t_1:t_2} = \chi \sum_{t=t_1}^{t_2} \varsigma_t \phi_t \quad (6)$$

where $\chi$ is trajectory weight. A cyclic trajectory will obtain a very large weight, e.g. $10^6$, to make its score discriminative from non-cyclic one.

**Main Procedure:** After introducing two critical components, main procedure of the proposed method is given in Alg. 1.

## 3. IMPLEMENTATION AND EXPERIMENTS

**Dataset and evaluated trackers:** The framework is referred as Multiple Trajectories of Single tracker (MTS). We evaluate it on CVPR2013 visual tracking benchmark [14], which calculates trackers' PR/SR, namely precision rate and success rate. Dataset consists of 51 test sequences with challenging factors, which are IV (illumination variation), SV (scale variation), OCC (occlusion), DEF (deformation), MB (motion blur), FM (fast motion), IPR (in-plane-rotation), OPR (out-of-plane rotation), OV (out-of-view), BC (background clutters), and LR (low resolution). The large and diverse dataset can bring a relatively unbiased evaluation. Evaluated tracking methods include ASLA [16], CSK [17], CXT [18], SCM [6], TLD [5], VTD [19], VTS [20], Struck [3] and KCF [21].

**Alg. 1.**

**Input:** frames $\{I_t\}$, $t \in [1,T]$, tracker amount $n$, interval length $\tau$, base tracker $\Gamma_{\text{base}}$.
**Output:** bounding box predictions $\{b_t\}$, $t \in [2,T]$.
1. $t \leftarrow 1$.
2. **do**
3.   Initialize $\{\Gamma_1, \Gamma_2, \ldots, \Gamma_n\}$ from $\Gamma_{\text{base}}$, $\mathbf{E} \leftarrow \{\Gamma_1, \Gamma_2, \ldots, \Gamma_n\}$.
4.   $t_1 \leftarrow t$, $t_2 \leftarrow t_1 + \tau$.
5.   **if** $t_2 < T$ **then** $t_2 \leftarrow T$.
6.   $\mathbf{E}$ tracks forward in interval $[t_1, t_2]$ through **Paced updates** and obtain forward trajectories $\{\overrightarrow{\mathbf{X}^i_{t_1:t_2}}\}$.
7.   $\mathbf{E}$ tracks backward in interval $[t_1, t_2]$ normally and obtain trajectory backward trajectories $\{\overleftarrow{\mathbf{X}^i_{t_2:t_1}}\}$.
8.   Select the best trajectory $\overrightarrow{\mathbf{X}^*_{t_1:t_2}}$ through **Trajectory Selection** with $\{\overrightarrow{\mathbf{X}^i_{t_1:t_2}}\}$ and $\{\overleftarrow{\mathbf{X}^i_{t_2:t_1}}\}$.
9.   Select $\Gamma^*$ which generates $\overrightarrow{\mathbf{X}^*_{t_1:t_2}}$.
10.   $[b_{t_1}, b_{t_2}] \leftarrow \overrightarrow{\mathbf{X}^*_{t_1:t_2}}$, $\Gamma_{\text{base}} \leftarrow \Gamma^*$.
11.   $t \leftarrow t_2 + 1$.
12. **while** $t < T$

**Parameter settings:** Two commonly recognized trackers are incorporated into MTS, Struck [3] and KCF [21], which are the typical methods employing SVM classifier and correlation filter correspondingly. In the test of MTS with Struck and KCF, tracker amount $n$ is 8 for both and interval $\tau$ is 10 and 20 correspondingly.

**General Evaluation:** We perform the one-pass evaluation (OPE) [14]. Fig. 3 shows the success plots and the precision plots. OPE scores of MTS with Struck exceed original

**Table 1**. Comparison of the PR/SR scores in the OPE method. Numbers in parenthesis in the first column refer to the numbers of sequences with the corresponding factors. Highest score is set bold in every test. The percentages in the right column of MTS+Struck and MTS+KCF is the increasing range comparing with original ones correspondingly.

|        | CXT         | SCM         | TLD         | VTD         | Struck      | KCF         | MTS+Struck  |              | MTS+KCF          |              |
|--------|-------------|-------------|-------------|-------------|-------------|-------------|-------------|--------------|------------------|--------------|
| IV(25) | 0.501/0.368 | 0.594/0.473 | 0.537/0.399 | 0.557/0.420 | 0.558/0.428 | 0.669/0.500 | 0.629/0.455 | 12.72%/6.31% | **0.742/0.544**  | 10.91%/8.80% |
| SV(28) | 0.550/0.398 | 0.672/0.528 | 0.606/0.432 | 0.597/0.408 | 0.639/0.434 | 0.717/0.539 | 0.694/0.454 | 8.61%/4.61%  | **0.766/0.558**  | 6.83%/3.53%  |
| OCC(29)| 0.491/0.380 | 0.640/0.496 | 0.563/0.412 | 0.545/0.406 | 0.564/0.421 | 0.797/0.573 | 0.696/0.496 | 23.40%/17.81%| **0.837/0.592**  | 5.02%/3.32%  |
| DEF(19)| 0.422/0.324 | 0.596/0.448 | 0.512/0.378 | 0.501/0.377 | 0.521/0.393 | 0.711/0.487 | 0.637/0.454 | 22.26%/15.52%| **0.722/0.496**  | 1.55%/1.85%  |
| MB(12) | 0.509/0.390 | 0.339/0.304 | 0.518/0.429 | 0.375/0.307 | 0.511/0.455 | 0.607/0.487 | 0.566/0.490 | 10.76%/7.69% | **0.715/0.538**  | 17.79%/10.47%|
| FM(17) | 0.515/0.404 | 0.333/0.300 | 0.551/0.435 | 0.352/0.300 | 0.604/0.479 | 0.613/0.477 | 0.648/0.518 | 7.28%/8.14%  | **0.685/0.515**  | 11.75%/7.97% |
| IPR(31)| 0.610/0.462 | 0.597/0.465 | 0.584/0.425 | 0.599/0.433 | 0.617/0.452 | 0.711/0.540 | 0.654/0.470 | 6.00%/3.98%  | **0.747/0.562**  | 5.06%/4.07%  |
| OPR(39)| 0.574/0.418 | 0.618/0.470 | 0.596/0.420 | 0.620/0.434 | 0.597/0.432 | 0.726/0.531 | 0.691/0.473 | 15.75%/9.49% | **0.774/0.563**  | 6.61%/6.03%  |
| OV(6)  | 0.510/0.427 | 0.429/0.361 | 0.576/0.457 | 0.462/0.446 | 0.539/0.459 | 0.735/0.602 | 0.589/0.507 | 9.28%/10.46% | **0.801/0.638**  | 8.98%/5.98%  |
| BC(21) | 0.443/0.348 | 0.578/0.461 | 0.428/0.356 | 0.571/0.430 | 0.585/0.471 | 0.571/0.476 | 0.626/0.477 | 7.01%/1.27%  | **0.663/0.502**  | 16.11%/5.46% |
| LR(4)  | 0.371/0.312 | 0.305/0.279 | 0.349/0.309 | 0.168/0.177 | 0.545/0.372 | 0.460/0.361 | **0.536**/0.391 | -1.65%/5.11% | 0.506/**0.401** | 10.00%/11.08%|
| Average| 0.575/0.426 | 0.649/0.499 | 0.608/0.437 | 0.576/0.416 | 0.656/0.474 | 0.744/0.546 | 0.727/0.505 | 10.82%/6.54% | **0.795/0.583**  | 6.85%/6.78%  |

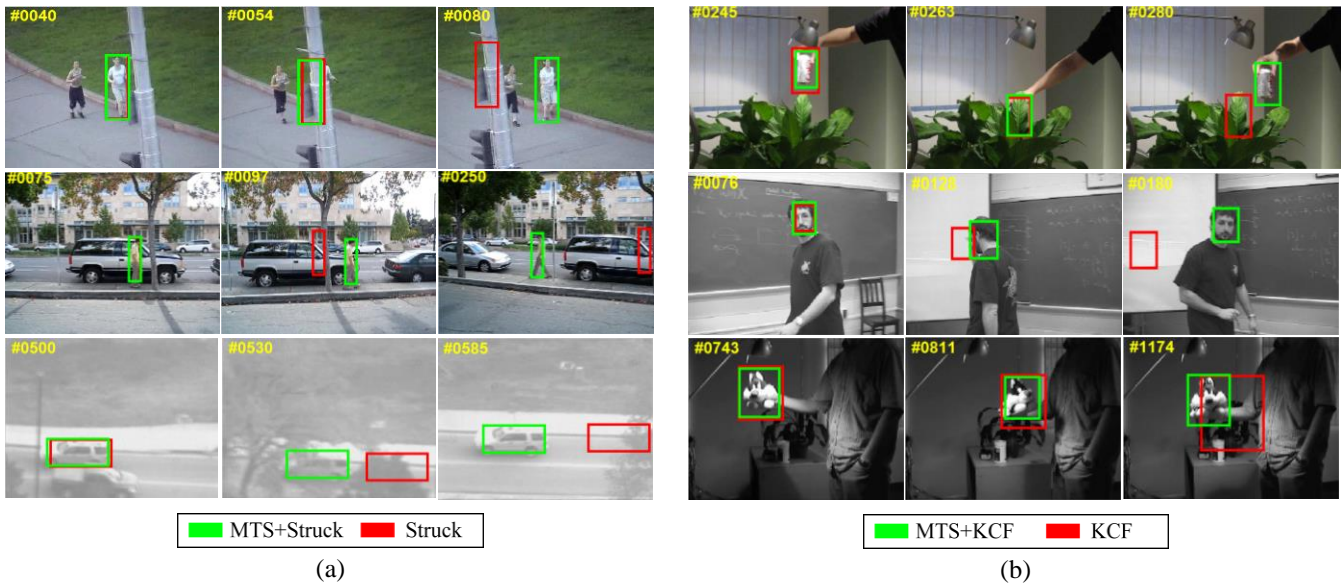

**Fig. 4**. Tracking screenshots of (a) MTS+Struck vs Struck, and (b) MTS+KCF vs KCF. Sequences are (a) jogging, david3, suv, (b) coke, freeman1, sylvester.

Struck's one from 0.656/0.474 to 0.727/0.505. MTS with KCF also gains a remarkable improvement comparing with original KCF from 0.744/0.546 to 0.795/0.583. Our method significantly outperforms all the trackers in [14].

**Factor Analysis:** We go further to compare with four trackers in Table 1 according to various challenging factors to demonstrate the advantages of the proposed framework. Since MTS helps a tracker utilize temporal context implicitly through trajectory selection and alleviate the influence of needless update information through paced updates, MTS achieves improvement in all challenging test, especially in illumination variation and motion blur, in which both trackers increase by more than 10% of PR/SR. In occlusion and deformation, MTS with Struck improves 23.40%/17.81% and 22.26%/15.52%. In motion blur and low resolution, MTS with KCF also improves 17.79%/10.47% and 10.00%/11.08%. Note that in low resolution, MTS with Struck obtain a slightly worse performance of -1.65%, but it can be considered as an experimental error since there are only 4 sequences in this scenario with limited lengths. In Fig. 4, the proposed framework can be seen to help original trackers avoid model drift and error scaling.

In summary, our methods can make up the drawbacks of trackers in various challenging scenarios.

## 4. CONCLUSION

In this paper, we proposed a tracking framework called MTS, with paced updates and trajectory selection, the cooperated tracker is enabled to implicitly take the advantage of temporal context and acquire the best situation through self-examination according to its forward and backward trajectories. Experimental results demonstrate the significant improvement made by our framework. Future research issues include a smarter strategy to pace the updates and a more precise trajectory analysis.

## 5. REFERENCES

[1] A. W. Smeulders, D. M. Chu, R. Cucchiara, S. Calderara, A. Dehghan, and M. Shah, "Visual tracking: an experimental survey," *TPAMI*, vol. 36, no. 7, pp. 1442–1468, 2014.

[2] N. Wang, J. Shi, D. Y. Yeung, and J. Jia, "Understanding and Diagnosing Visual Tracking Systems," *ArXiv150406055 Cs*, Apr. 2015.

[3] S. Hare, A. Saffari, and P. H. Torr, "Struck: Structured output tracking with kernels," in *Computer Vision (ICCV), 2011 IEEE International Conference on*, 2011, pp. 263–270.

[4] B. Babenko, M.-H. Yang, and S. Belongie, "Robust object tracking with online multiple instance learning," *TPAMI*, vol. 33, no. 8, pp. 1619–1632, 2011.

[5] Z. Kalal, K. Mikolajczyk, and J. Matas, "Tracking-learning-detection," *TPAMI*, vol. 34, no. 7, pp. 1409–1422, 2012.

[6] W. Zhong, H. Lu, and M.-H. Yang, "Robust object tracking via sparsity-based collaborative model," in *Computer Vision and Pattern Recognition (CVPR), 2012 IEEE Conference on*, 2012, pp. 1838–1845.

[7] W. Hu, X. Li, W. Luo, X. Zhang, S. Maybank, and Z. Zhang, "Single and multiple object tracking using log-euclidean riemannian subspace and block-division appearance model," *TPAMI*, vol. 34, no. 12, pp. 2420–2440, 2012.

[8] C. Ma, X. Yang, C. Zhang, and M.-H. Yang, "Long-Term Correlation Tracking," in *Computer Vision and Pattern Recognition (CVPR), 2015 IEEE Conference on*, 2015, pp. 5388–5396.

[9] H. U. Kim, D. Y. Lee, J. Y. Sim, and C. S. Kim, "SOWP: Spatially Ordered and Weighted Patch Descriptor for Visual Tracking," in *Computer Vision (ICCV), 2015 IEEE International Conference on*, 2015, pp. 3011–3019.

[10] H. Possegger, T. Mauthner, and H. Bischof, "In Defense of Color-based Model-free Tracking," in *Proceedings of the IEEE Conference on Computer Vision and Pattern Recognition*, 2015, pp. 2113–2120.

[11] N. Wang, S. Li, A. Gupta, and D.-Y. Yeung, "Transferring Rich Feature Hierarchies for Robust Visual Tracking," *ArXiv Prepr. ArXiv150104587*, 2015.

[12] Z. Hong, Z. Chen, C. Wang, X. Mei, D. Prokhorov, and D. Tao, "MUlti-Store Tracker (MUSTer): A cognitive psychology inspired approach to object tracking," in *2015 IEEE Conference on Computer Vision and Pattern Recognition (CVPR)*, 2015, pp. 749–758.

[13] J. Zhang, S. Ma, and S. Sclaroff, "MEEM: Robust tracking via multiple experts using entropy minimization," in *Computer Vision–ECCV 2014*, Springer, 2014, pp. 188–203.

[14] Y. Wu, J. Lim, and M.-H. Yang, "Online object tracking: A benchmark," in *Computer Vision and Pattern Recognition (CVPR), 2013 IEEE Conference on*, 2013, pp. 2411–2418.

[15] D. Y. Lee, J. Y. Sim, and C.-S. Kim, "Multihypothesis Trajectory Analysis for Robust Visual Tracking," in *Computer Vision and Pattern Recognition (CVPR), 2015 IEEE Conference on*, 2015, pp. 5088–5096.

[16] X. Jia, H. Lu, and M.-H. Yang, "Visual tracking via adaptive structural local sparse appearance model," in *Computer Vision and Pattern Recognition (CVPR), 2012 IEEE Conference on*, 2012, pp. 1822–1829.

[17] J. F. Henriques, R. Caseiro, P. Martins, and J. Batista, "Exploiting the Circulant Structure of Tracking-by-detection with Kernels," in *Computer Vision–ECCV 2012*, 2012.

[18] T. B. Dinh, N. Vo, and G. Medioni, "Context tracker: Exploring supporters and distracters in unconstrained environments," in *Computer Vision and Pattern Recognition (CVPR), 2011 IEEE Conference on*, 2011, pp. 1177–1184.

[19] J. Kwon and K. M. Lee, "Visual tracking decomposition," in *Computer Vision and Pattern Recognition (CVPR), 2010 IEEE Conference on*, 2010, pp. 1269–1276.

[20] J. Kwon and K. M. Lee, "Tracking by sampling trackers," in *Computer Vision (ICCV), 2011 IEEE International Conference on*, 2011, pp. 1195–1202.

[21] J. F. Henriques, R. Caseiro, P. Martins, and J. Batista, "High-Speed Tracking with Kernelized Correlation Filters," *TPAMI*, 2014.